# MolSets: Molecular Graph Deep Sets Learning for Mixture Property Modeling

Hengrui Zhang,[1] Jie Chen,[1] James M. Rondinelli,[2,*] Wei Chen[1,*]

[1]Department of Mechanical Engineering, Northwestern University, Evanston, IL 60208, USA

[2]Department of Materials Science and Engineering, Northwestern University, Evanston, IL 60208, USA

*Emails: jrondinelli@northwestern.edu (J.M.R.), weichen@northwestern.edu (W.C.)

**ABSTRACT**

Recent advances in machine learning (ML) have expedited materials discovery and design. One significant challenge faced in ML for materials is the expansive combinatorial space of potential materials formed by diverse constituents and their flexible configurations. This complexity is particularly evident in molecular mixtures, a frequently explored space for materials such as battery electrolytes. Owing to the complex structures of molecules and the sequence-independent nature of mixtures, conventional ML methods have difficulties in modeling such systems. Here we present MolSets, a specialized ML model for molecular mixtures. Representing individual molecules as graphs and their mixture as a set, MolSets leverages a graph neural network and the deep sets architecture to extract information at the molecule level and aggregate it at the mixture level, thus addressing local complexity while retaining global flexibility. We demonstrate the efficacy of MolSets in predicting the conductivity of lithium battery electrolytes and highlight its benefits in virtual screening of the combinatorial chemical space.

**INTRODUCTION**

The design of materials and molecules requires understanding the structure–property relationship in a broad chemical space. Navigating the chemical space is challenging because the *combinatorial complexity*, originating from the diversity of constituents (e.g., atoms) and various configurations of the constituents (e.g., atomic arrangements), forms an expansive space with the number of candidates far exceeding the capability of experiments or computational simulation[1]. Despite the challenges, growing amounts of data have been collected experimentally and computationally in this pursuit. Machine learning (ML) methods can harness this data and efficiently establish structure–property relationships[2]. ML has shown potential in predictive modeling, virtual screening, as well as accelerating the design of novel materials and molecules[3-5].

For many materials systems of practical importance, however, the combinatorial complexity occurs not only at one level but multiple levels. An example is *molecular mixtures*, a chemical space frequently explored in search of candidates for electrolytes[6], coolants, and fuels[7], among other applications. Previously, ML models have been applied to physical properties in small subspaces[8, 9], such as a binary liquid. The applicability of current ML models in predicting more complex functions within a broader space of molecular mixtures is limited by the multilevel combinatorial complexity: The arrangement of various atoms produces complexity at the molecular level, whereas the mixing of different molecules brings added complexity at the mixture level. The key challenge lies in *representation*[10], i.e., converting the structure into a digital format that the ML model can access and use. For accurately learning structure–property relationships of molecular mixtures, it is crucial to find a meaningful representation that both (1) captures the relevant physical and chemical information and (2) reflects the similarity between datapoints.

Specifically, the challenge of representing mixtures and predicting their properties is three-fold. First, for an individual molecule, the property is determined by its complex chemistry and geometry, which should be encoded in the representation and exposed to the model. The atomic properties and molecular geometry form the essential chemistry to be captured. Owing to its versatility in encoding this information, graphs have been widely adopted as a representation of molecules. Using the graph representation, graph neural networks (GNNs) have demonstrated efficacy in molecular modeling and design applications[11-14]. Second, between molecules, there are interactions that influence their structures and behaviors, making the mixture property non-additive, which requires flexible methods that do not a priori assume a form of constituent interactions. Last, a random mixture is sequence-independent, i.e., (30% A, 70% B) is the same as (70% B, 30% A). In conventional ML models which take vector inputs, the two ways of representing the same mixture will be viewed differently, which fails to reflect the similarity between datapoints and leads to incorrect outputs. To address this, the representation should possess *permutation invariance*, as "sets" in mathematics, and ML model architectures such as Deep Sets[15, 16] were developed for attaining permutation-invariant modeling of sets.

In this work, we represent a molecular mixture as a set of molecular graphs and propose the MolSets ML model for predicting mixture properties from this representation. MolSets leverages (1) GNN to extract information from molecular chemistry and geometry, (2) the attention mechanism[17] to learn the relative importance and interaction of constituents, and (3) the Deep Sets architecture to ensure permutation invariance. We demonstrate the predictive power and interpretability of MolSets in modeling the electrical conductivity of molecular mixtures to facilitate the virtual screening of lithium battery electrolytes.

**PROBLEM FORMULATION**

The electrolyte is an essential component of various types of lithium batteries[18, 19], and a major family of electrolytes are solid, liquid, or gel mixtures of molecules or polymers[20-22]. Broadly, we may view these all as molecular mixtures, with a single-component molecule as a special case thereof. The design of electrolytes involves a range of metrics to be considered, including property metrics such as electrical conductivity, performance metrics such as Coulombic efficiency (CE), as well as safety and sustainability metrics. To demonstrate our proposed MolSets model, we choose the room temperature (298 K) electrical conductivity as a target property to be predicted from the molecular mixture formulation (constituents and their weight fractions). Figure 1 illustrates the problem formulation. Predicting conductivity serves as a preliminary step of the virtual screening of electrolytes, and the MolSets model is generally applicable to other properties or performance metrics.

We retrieve a dataset curated by Bradford et al.[6] from previously published literature. The dataset reports experimentally measured conductivity of polymer or molecular mixtures, each consisting of up to 4 types of molecules, together with Li$^+$ salts and inorganic additives. To avoid the missing values, we select a subset of distinct mixtures, each of which is formulated by 1–4 different molecules and one salt, as a testbed for MolSets. Note that not all mixtures have conductivity reported at 298 K in the dataset, but they all have conductivities reported at multiple different temperatures. We use the reported ionic conductivities $\sigma$ to perform a linear fit based on the Arrhenius transport

$$\log \sigma = -\frac{E_a}{RT} + \log \sigma_\infty = k \cdot \frac{1}{T} + b \qquad (1)$$

where the temperature dependence $k$ is related to the activation energy $E_a$ and ideal gas constant $R$. With the linear fit, we calculate an inferred 298 K conductivity for the mixtures without a reported value in the dataset.

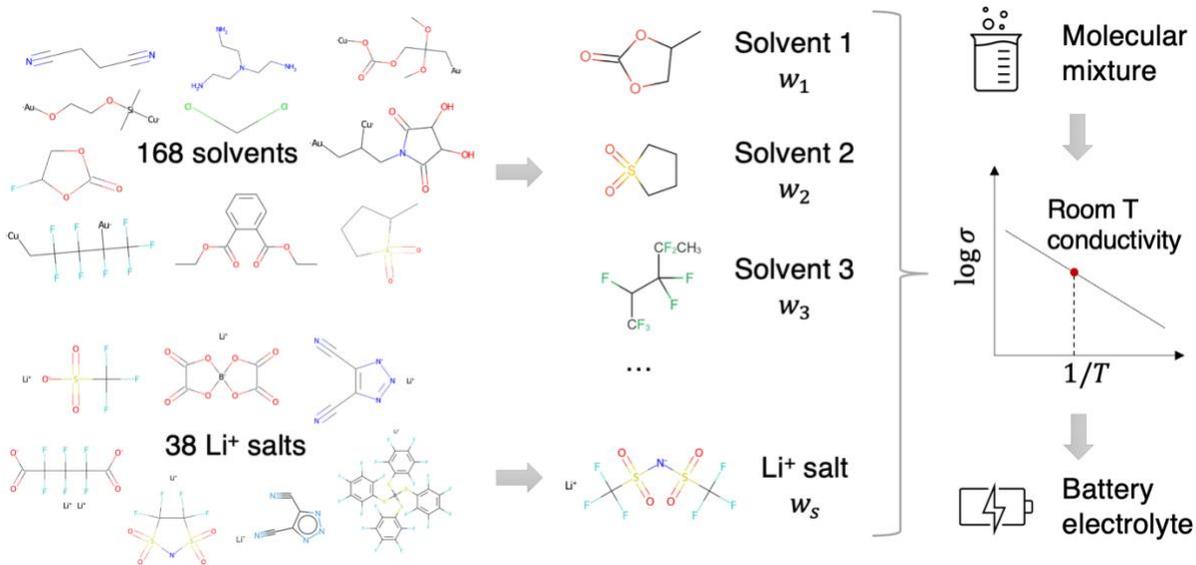

Figure 1. Overview of the electrolyte conductivity prediction problem. We consider the molecular mixtures consisting of 1–4 solvents and 1 Li$^+$ salt in a candidate space of 168 types of solvents and 38 types of salts. Structures of some representative constituents are visualized. The solvents include both small molecules and polymers; for polymers, we show the structure of their monomer, where "Cu" and "Au" are placeholders to indicate the connecting sites of the monomer. (Cu and Au are not included in the molecular graph.) Our model is used to predict the room temperature conductivity of a molecular mixture as an indicator of its potential as a battery electrolyte.

**MODEL DEVELOPMENT**

Deep Sets Learning

A mixture with arbitrary constituents can be represented as a set $X = \{x_1, x_2, \ldots, x_m\}$, where each $x$ is a distinct constituent, and the number of constituents $m$ is not fixed. To model a mixture property $y = f(X)$, permutation invariance should be ensured:

$$f(\{x_1, \ldots, x_m\}) = f(\{x_{\pi(1)}, \ldots, x_{\pi(m)}\}) \qquad (2)$$

for any permutation of sequence $\pi$. The Deep Sets architecture defines the sufficient and necessary principle for building permutation invariant models:

$$f(X) = \rho\left(\sum_{x \in X} \phi(x)\right) \tag{3}$$

or equivalently, $f(X) = \rho(\oplus_{x \in X}\{\phi(x)\})$, where $\phi(\cdot)$ and $\rho(\cdot)$ are appropriate transformations, and $\oplus$ is any permutation invariant aggregating operation.

A unique feature of molecular mixtures is that the set representation becomes $X = \{(x_i, w_i)\}_{i=1}^{m}$, where each molecule $x$ has a weight fraction $w$ in the mixture. The permutation invariant model is modified accordingly:

$$f(X) = \rho\left(\oplus_{(x,w) \in X}\{\phi(x), w\}\right) \tag{4}$$

In our implementation, the model consists of three components: (1) an "embedding" module $\phi$ that learns a latent representation for each molecule; (2) an "aggregation" module $\oplus$ that combines representations of individual molecules into a latent representation for the mixture; and (3) a "transformation" module $\rho$ that maps the mixture representation to the target function. Figure 2a illustrates the overall model architecture.

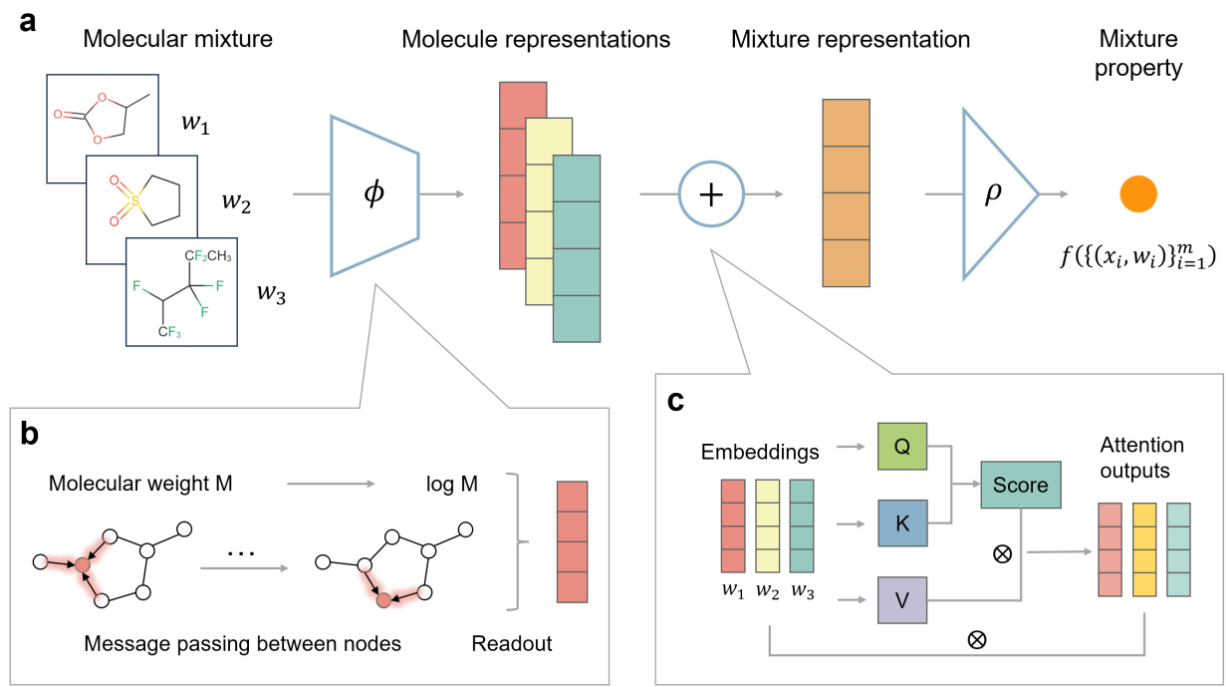

Figure 2. (a) The architecture of MolSets. (b) The "embedding" module $\phi$ is a graph neural network, which performs iterative message passing between nodes and reads out all node features to form a representation of the molecule. (c) The "aggregation" module $\oplus$ first uses the attention mechanism to adjust the molecular representations according to their importance, and then performs a weighted sum to form a mixture

representation. The "transformation" module $\rho$ is a multilayer perceptron composed of several fully connected layers, which maps the mixture representation to the target property.

## Molecular Graph Neural Network

We convert a molecule into the graph data structure, $G = (V, E)$, where the nodes $V$ represent heavy atoms (any element except H) in the molecule, and edges $E$ represent the bonds between them. Every node is associated with a list of features. To make it generalizable, we use only the element type and easily obtainable atomic descriptors for an atom: atomic mass, formal charge, electronegativity, van der Waals radius, and the number of H atoms connected to it. Every bond is associated with one feature, the bond type (single, aromatic, double, etc.). The atomic and bond descriptors are obtained using open-source cheminformatics software RDKit[23] and pymatgen[24]. In addition, to distinguish between small molecules and polymers, we associate every graph with the logarithmic molecular weight $\log M$ as a graph-level feature.

We use a graph neural network (GNN) to learn the molecular representations. As Figure 2b shows, for an input graph, the GNN performs message passing between connected nodes using a "graph convolution" operator:

$$v_i' \leftarrow MP(v_i, v_j, e_{ij})_{j \in N(i)} \text{ for } v_i \in V \tag{5}$$

In one convolutional layer, the feature vector of every node is updated using a "message" derived from features of its neighbor nodes and edges between them. How the message is derived varies depending on the type of convolution operator. After several convolutional layers, the global mean value vector of all node features is calculated. This vector is concatenated with $\log M$, and goes into a fully connected layer, which generates a vector representation of the graph.

## Aggregation with Attention Mechanism

The $\phi$ module maps every molecular graph $x$ to a vector representation $z$. To aggregate the individual molecular representations and weight fractions $\{(z_i, w_i)\}_{i=1}^m$, an intuitive way is to perform a weighted summation. However, a constituent's contribution to the mixture property is not solely, linearly dependent on its weight fraction. We use the attention mechanism[17] to more accurately capture the constituents' contributions. For each molecular representation $z$, a "query" vector $q$ is calculated by $q = z \cdot W^Q$, with learnable parameters $W^Q$, and similarly for two other vectors, "key" $k$ and "value" $v$. Then, a "score" that reflects the molecule's importance is derived from $q$ and $k$, from which we get an updated molecular representation

$$z' = \text{score} \cdot v = \text{softmax}\left(\frac{q \cdot k^{\text{T}}}{\sqrt{d_k}}\right) \cdot v \qquad (6)$$

where $d_k$ is the length of $k$, and $\text{softmax}(\cdot)$ is a vector function $\boldsymbol{\sigma}(\mathbf{x})_i = \frac{\exp x_j}{\sum_{j=1}^{n} \exp x_j}$ for any vector $\mathbf{x} = [x_1, \ldots, x_n]$. The updated representation does not change in dimension, but its value is modulated according to its importance. In addition, permutation invariance is maintained as the attention mechanism operates on each molecule separately with shared parameters.

Afterward, from the updated representations of molecules, the representation of the mixture is computed as a weighted sum: $Z = \sum_{i=1}^{m} w_i z_i'$. These form the "aggregation" module $\oplus$, which maps an arbitrary number of molecular representation vectors into a single mixture representation, as shown in Figure 2c. The aggregation formulation accounts for the weight fractions of constituent molecules, as well as the nonadditive or nonlinear contributions of constituents on the mixture property.

## MolSets Model Architecture

Based on the Deep Sets guidelines (Equation 4) and the two modules described above, we propose the MolSets model architecture, illustrated in Figure 2. A molecular mixture is input as a set of graphs. The graphs are mapped by a GNN-based embedding module $\phi$ to representation vectors, which are then aggregated to a mixture representation vector by the $\oplus$ module. Finally, the transformation module $\rho$, a few layers of fully connected neural network, maps the mixture representation to the output, i.e., the target property of the mixture. Table 1 lists detailed specifications of the MolSets model.

Table 1. The implementation of three modules of MolSets and hyperparameters that could impact the model architecture.

| Module | Implementation | Key hyperparameters |
|---|---|---|
| Embedding $\phi$ | Graph convolutional layers | Convolution operator type; Number and dimensions of layers |
| | Global mean pooling | |
| | Fully connected layer | Dimension of representation |
| Aggregation $\oplus$ | Attention layer | Dimension of $q, k, v$ |
| | Weighted sum | |
| Transformation $\rho$ | Fully connected layers | Number of layers; Dimensions of hidden layers |

## RESULTS AND DISCUSSIONS

### Model Testing

As a demonstration of MolSets, we perform tests on the electrolyte conductivity prediction task described in Problem Formulation. Since an electrolyte contains two different types of molecules, solvents and salts, they can be treated either together or separately in the model. Here, we adopt a more general setting: constituents can be grouped into several categories, and a molecular mixture can be considered as mixtures within each category. In the case where no grouping is applicable, this reduces to the "all together" setting.

Following the setting, we adjust MolSets to a dual-pathway architecture to be compatible with the data: two $\phi$ modules are used, learning representations for solvents and salts, respectively. As each electrolyte consists of multiple solvents and only one salt in the dataset we use, the learned representations for solvents are aggregated into solvent mixture representation using $\oplus$, while that for the salt needs no aggregation and is by itself the "salt mixture representation". If multiple salts are considered, the salt mixture representation can be learned in the same way as solvents using $\phi$ and $\oplus$. Then, the solvent mixture representation, salt representation, and salt molality are concatenated and passed to $\rho$.

In the following part, we show results of predicting electrolyte conductivity using MolSets and other methods. These tests serve three purposes: (1) As MolSets is a generic architecture that can work with various types of GNNs and has several hyperparameters, we explore the effect of some key configurations on its performance. (2) Benchmarks that demonstrate MolSets' advantage over existing models. (3) Ablation tests that investigate the importance of different parts of MolSets. We use two metrics for quantifying model performance: (1) Pearson correlation coefficient $r_p$, which measures the linear correlation between target and predicted values. It is the square root of another commonly used metric, coefficient of determination $R^2$. (2) Spearman rank correlation coefficient $r_s$, which measures how well the predicted values are ranked correctly as the targets. It is important for materials modeling, especially for materials properties to be optimized, as correct ranking can guide the search towards superior candidates. Since the combinations of models and configurations lead to many tests, cross-validation becomes time-consuming, hence, we randomly split the dataset into training, validation, and testing datasets in a 3:1:1 ratio. For every different model and configuration, hyperparameter tuning is performed using the training and validation data, and final performance metrics are assessed on the testing data.

### Benchmark and Ablation Test

A major customizable configuration of MolSets is the type of GNN used as the $\phi$ module. An abundance of GNNs have been developed, and their main difference lies in the "graph convolution" operation, i.e., how messages are composed and passed between nodes. Out of many available, we choose four commonly used general-purpose graph convolution operators, GraphConv, SAGEConv, GCNConv, and GATConv, as well as a directed message passing scheme (DMPNN) specially designed for molecules[25]. Integrating each convolution operator to MolSets, we tune the key hyperparameters that have an impact on the model architecture using the validation dataset and report the testing performances $r_p$ and $r_s$. The models are implemented using the PyTorch Geometric[26] library, and more details are presented in the supplementary information (SI).

The first few rows of Table 2 list the performance of MolSets integrated with different graph convolution operators. Using GraphConv and SAGEConv, both generic and lightweight convolution operators, MolSets attain high regression and ranking accuracies on the testing data. DMPNN also shows high testing accuracy; though specially designed for molecules, the performance of DMPNN is not superior possibly because its architecture is not tailored for working with the $\oplus$ and $\rho$ modules as a part of MolSets.

Table 2. Model configurations and their performances. We list whether the models strictly preserve permutation invariance by design, and their $r_p$ and $r_s$ metrics on the testing dataset.

| Model | Configuration | Permutation invariance | Pearson $r_p$ | Spearman $r_s$ |
|---|---|---|---|---|
| MolSets | GraphConv | Yes | 0.905 | 0.907 |
|  | SAGEConv |  | 0.898 | 0.905 |
|  | GCNConv |  | 0.874 | 0.872 |
|  | GATConv |  | 0.877 | 0.884 |
|  | DMPNN |  | 0.892 | 0.897 |
| GNN | Weighted Sum | Yes | 0.874 | 0.882 |
|  | Concatenate | No | 0.853 | 0.869 |
| GBDT | LightGBM | No | 0.864 | 0.876 |

Next, we train two different GNN-based deep learning models to predict mixture properties from the constituents' molecular graphs. (1) Replace the $\oplus$ aggregation module of MolSets with a simple weighted summation. This model still satisfies Equation 3 and is thus permutation invariant. (2) Instead of aggregation, the molecular representation vectors learned by a GNN $\phi$ are concatenated and padded with zeroes, together with solvents' weight fractions and salt molality, to form a mixture representation vector, which then goes through $\rho$. This model no longer preserves permutation invariance and needs the maximum number of constituents to be specified. We choose GraphConv as the convolution operator in the GNN and, to control variables, keep other hyperparameters the same as used in the MolSets model with GraphConv.

As another benchmark, we retrieve a list of numerical molecular descriptors and predict the conductivity based on these descriptors using gradient boosting. 208 descriptors are calculated for every molecule using RDKit. Then, for every mixture, the descriptors of its constituents (4 solvents and 1 salt) together with their logarithmic molecular weights and weight fractions are concatenated to form a feature vector. For the mixtures containing less than 4 solvents, the positions corresponding to missing solvents are filled with zeros. The dimensions containing abnormal values (NaN or infinity) or have no variance across data points are removed. LightGBM[27], an accurate and efficient implementation of the gradient boosting decision tree (GBDT) algorithm, is employed to fit a model using the training and validation sets, and its performance on the testing set is shown in Table 2. Like the GNN with concatenation, this method is also not permutation invariant and restricted to a prespecified number of constituents.

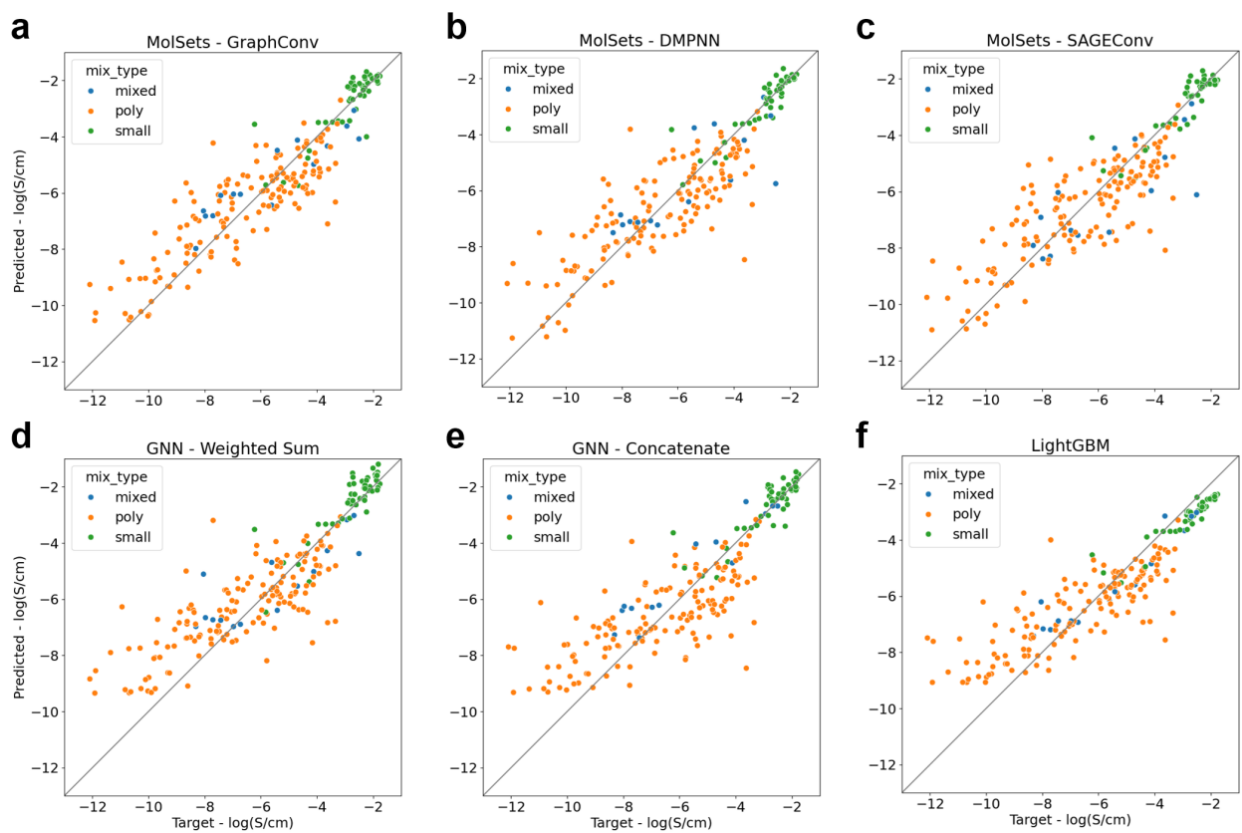

Figure 3. Regression plots showing the target values (horizontal axis) and predicted values (vertical axis) of logarithmic conductivity. (a–c) MolSets using different graph convolution operators. (d–e) GNN models using the GraphConv operator and different treatments of molecular representations learned from graphs. (f) LightGBM model trained on molecular descriptors. "Mix_type" denotes whether a data point is a mixture of molecules ("small"), polymers ("poly"), or both ("mixed").

Table 2 lists the performance metrics and whether permutation invariance is preserved for all models. In Figure 3, we show the true values versus model-predicted values of logarithm conductivity for the testing data, with colors differentiating different types of mixtures. In general, MolSets' predictions show less deviation from the true values; moreover, MolSets models show lower errors at the high-end and low-end of conductivity (points far from the center), which favors the discovery of exceptional materials[28].

## Analyses and Interpretation

The fundamental uniqueness of MolSets, compared to other ML models, is that it treats a mixture as a permutation-invariant set of constituents whose contributions are not simply governed by their weight fractions. The advantage of this assumption is demonstrated by the better predictive performance. To interpret why this assumption can lead to better performance, we investigate two questions: (1) How is a mixture different from the weighted summation of its constituents? (2) What happens if a mixture model is not permutation invariant?

For (1), we probe the representation space learned by the $\phi$ and $\oplus$ modules of MolSets. Every molecular mixture, input to MolSets with GraphConv as a set of graphs, is mapped to a 32-dimensional representation vector. Note that a "mixture" with only one constituent can be viewed as a representation of that constituent. Choosing three small molecules, together with one binary mixture and one ternary mixture among them, we investigate their locations in the representation space. We use t-distributed Stochastic Neighbor Embedding (t-SNE)[29] to reduce the 32-dimensional vectors into 2 dimensions, and visualize the locations of constituents, mixtures (marked by text), and weighted summations of constituents (indicated by arrows) in the representation space in Figure 4a. The significant deviation of the mixture representation from the weighted summations (dashed lines) suggests that the mixture is not formed as a linear combination of its constituents together, which necessitates a specialized model like MolSets.

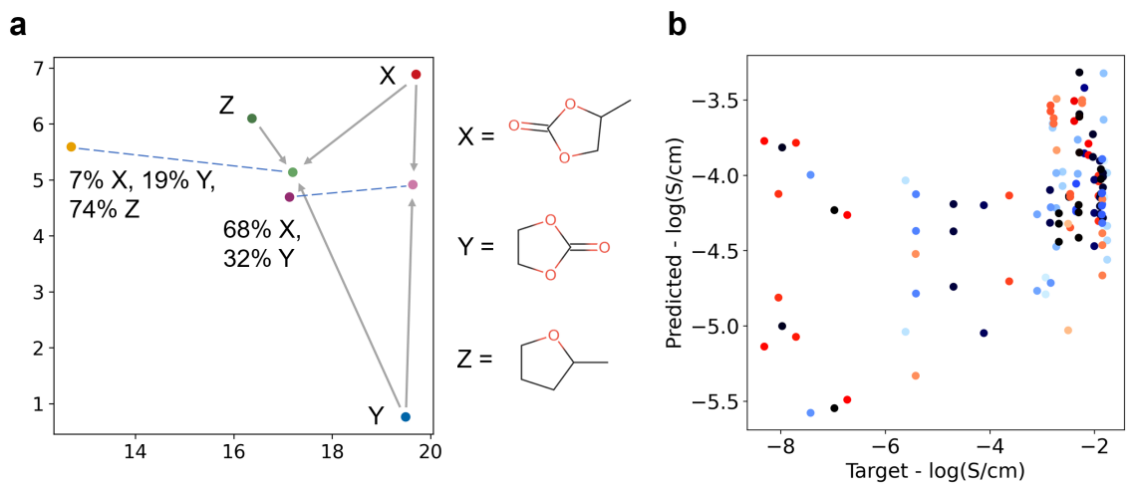

Figure 4. (a) Reduced-dimensional visualization of three molecules (shown on the right) and their binary/ternary mixtures in the learned representation space. The arrows indicate the weighted summation of constituents' representations, whose deviation from the learned representations of mixtures is indicated by dashed lines. (b) The target values (horizontal axis) and GNN-predicted values (vertical axis) on the permuted dataset. Colors are assigned according to target values to help distinguish different mixtures.

For (2), we test the performance of the non-permutation invariant GNN model (concatenating molecular representations) on a dataset containing mixtures represented in different sequences. From the testing dataset, we select all the mixtures that contain more than one solvent and create a new dataset where the solvents in every mixture appear in different orders. For each permuted datapoint, the order of solvents, molecular weights, and weight fractions are permuted in the same way, hence, it represents the same mixture, and the target property is unchanged. Figure 4b shows the GNN model's predictions on this permuted dataset: it gives substantially different predictions for the same mixture represented in different orders. Moreover, the model's predictions show systematic deviation from the target values on these mixtures containing more than one constituent. These demonstrate the advantage of permutation invariance in mixture modeling.

## Virtual Screening

Finally, we employ the MolSets model with the best configurations found in tests to perform virtual screening. We train the model on the whole dataset (70% for training, 30% for validation), and use it to predict the room temperature conductivity for a large set of candidates among the small molecules and $Li^+$ salts that appear in at least 3 mixtures in the dataset. We consider all equal-weight binary mixtures among 28 types of small molecules, combining with 30 types of salts (1 mol/kg), totaling 11340 candidates. The predicted conductivities of all candidates will be available at Dryad. Out of the candidates, the top-performing ones are listed in Table S3. Fixing the type and molality of salt, we compare the equal-weight binary mixtures with their constituents and find that a binary mixture can lead to higher conductivity than either of its constituent molecules (Table S4). This indicates a nontrivial improvement in the properties of mixtures, which resembles the *bowing effect* observed in alloys[30], and provides the potential of designing electrolytes with superior properties to single constituents within the broader chemical space of mixtures.

A limitation of the current model is that it does not take salt solubility into account. Since the dataset is curated from experimental literature, where salt molality does not exceed solubility, the model trained thereon cannot generalize to mixtures containing higher salt concentrations. Nonetheless, by rationally choosing salt molality as a constraint, this issue can be prevented when applying MolSets to electrolyte screening.

## Summary and Outlook

In summary, we presented MolSets, a machine learning model architecture for molecular mixtures that captures the chemistry and geometry of molecules while preserving the permutation invariance nature of mixtures. Using the conductivity of electrolytes as a testbed, we demonstrated the accuracy and robustness of MolSets, and investigated the nontrivial characteristics of mixtures compared to the combination of constituents.

An accurate yet efficient predictive model like MolSets can facilitate the virtual screening of promising materials in the vast combinatorial space of molecular mixtures. As an initial step, we use MolSets to predict the conductivity of over 10000 mixtures based on the available data. However, the limited availability of data poses a key challenge. Although a molecular property database has been released recently[31], there is no comprehensive data resource for mixtures. A future direction can be constructing a platform where researchers can access MolSets-predicted properties of mixtures and upload experimentally measured values. With growing amounts of data and the MolSets model, such a platform can offer accurate estimates of molecular mixture properties, as AlphaFold[32] offers for proteins and matterverse.ai[33] for crystals.

## ADDITIONAL INFORMATION

### Author Contributions

H.Z. conceived the project, developed the methods, conducted the tests, and drafted the manuscript. J.C. contributed to method development and analysis. J.M.R. and W.C. supervised the project. All authors reviewed and revised the manuscript.

### Data and Code Availability

Raw data is retrieved from Ref.[6] The code and processed data will be made publicly available at https://github.com/Henrium/MolSets. The newly generated data by virtual screening will be available at Dryad (link provided in the GitHub repository).

### Acknowledgments

This work was supported in part by the National Science Foundation (NSF) under grant numbers DMR-2324173 and DMR-2219489. H.Z. was supported by the Ryan Graduate Fellowship. The authors thank Tianxing Lai, Liwei Wang, Yaxin Cui, Akash Pandey, Xiaoyu Xie, Daniel Apley, Guilherme Missaka, Jeffrey Lopez, Yuxin Chang, and Edward Sargent for helpful discussions.

# Supplementary Information for

## MolSets: Molecular Graph Deep Sets Learning for Mixture Property Modeling

Hengrui Zhang,[1] Jie Chen,[1] James M. Rondinelli,[2,*] Wei Chen[1,*]

[1]Department of Mechanical Engineering, Northwestern University, Evanston, IL 60208, USA

[2]Department of Materials Science and Engineering, Northwestern University, Evanston, IL 60208, USA

*Emails: jrondinelli@northwestern.edu (J.M.R.), weichen@northwestern.edu (W.C.)

## DATASET PREPARATION

### Molecular Graph Formulation

A molecule is converted to a graph with its heavy atoms (any element other than H) as nodes and bonds between them as edges. The node features are 13-dimensional, with the first 7 dimensions being one-hot encoding of elements B, C, N, O, F, S, and Cl. If the atom is one of these elements, the corresponding dimension is 1 and others are 0; otherwise, these 7 dimensions are all 0. The remaining 6 dimensions are atomic number, atomic mass (in Dalton), formal charge (in elementary charge $e$), electronegativity (in Pauling scale), van der Waals radius (in Angstroms), and number of H atoms connected to the atom. Each node is associated with one feature, a numerical representation of the bond type: 1, 2, 3 for single, double, and triple bonds, respectively, and 1.5 for aromatic bonds.

We use a dataset involving both small molecules and polymers due to limited available data for mixture properties. We represent a polymer with the molecular graph of its monomer; as a monomer is linked to another in the real polymer, we add C atoms to its connecting sites in the graph. In addition, we include the molecular weight $M$ in the input so that polymers can be distinguished from small molecules. In the dataset, some polymers have the weight average molecular weight $M_w$ reported and some have the number average molecular weight $M_n$. We use the available one as $M$ and use $M_n$ if both are available.

## METHODS AND MODEL

### Graph Convolution

A graph is formulated as $G = (V, E)$, with nodes $V = \{v_i\}_{i=1}^n$ and edges $E = \{e_{ij}\}$. Each node $v_i$ is associated with a node feature vector $x_i$, and in our formulation, each edge is associated with a scalar feature, denoted $e_{ij}$ for simplicity. Graph convolution iteratively updates node feature vectors. A graph convolution operator defines (1) how a "message" is composed of a node $v_i$ neighbor nodes $j \in N(i)$ and edges $e_{ij}$ between them, and (2) how the node's feature is updated based on the message.

Table S1 shows the graph convolution operators used in this paper and their message passing schemes. In the formulas, **W**s denote learnable weight matrices; GCNConv utilizes the quantity $d_i = 1 + \sum_{j \in N(i)} e_{ij}$; GATConv involves pairwise attention coefficients $\alpha_{ij}$. Weight matrices are learned from and used for all nodes. More details are provided in the online documentation of PyTorch Geometric[1].

Table S1. Definitions of graph convolution operators.

| Convolution Operator | Message Passing Scheme |
|---|---|
| SAGEConv[2] | $x_i' = \mathbf{W}_1 x_i + \mathbf{W}_2 \cdot \text{mean}_{j \in N(i)} x_j$ |
| GraphConv[3] | $x_i' = \mathbf{W}_1 x_i + \mathbf{W}_2 \cdot \sum_{j \in N(i)} e_{ij} \cdot x_j$ |
| GCNConv[4] | $x_i' = \mathbf{W}^T \cdot \sum_{j \in N(i) \cup \{i\}} \frac{e_{ij}}{\sqrt{d_i d_j}} x_j$ |
| GATConv[5] | $x_i' = \alpha_{ij} \mathbf{W}_1 x_i + \sum_{j \in N(i)} \alpha_{ij} \mathbf{W}_2 x_j$ |

### Model Training and Hyperparameter Tuning

The model is trained using an AdamW optimizer[6] with an initial learning rate of 0.001 and a weight decay coefficient of 0.0001 for regularization. The learning rate is controlled using a scheduler, which reduces it by a factor of 2 upon 10 epochs of no improvement in validation loss. To further prevent overfitting, an early stopping rule is adopted in training: if validation loss has not improved for 20 successive epochs, the training process is terminated and the model parameters at the epoch that displayed the lowest validation loss are taken as the final model parameters.

Implementing MolSets with each graph convolution operator, we tune the hyperparameters that have an impact on model architecture, which include (1) the number of convolution layers, (2) the dimension of hidden layers, (3) the dimension of learned molecular representation, and (4) dimension of attention. The tuning is conducted using grid search over a discrete set of values for each hyperparameter and tracked using the Weights & Biases platform[7]. Table S2 lists the hyperparameters found optimal for MolSets with different convolution operators.

Table S2. Tuned hyperparameters for MolSets using each convolution operator.

| Convolution | Convolution layers | Hidden dimension | Representation dimension | Attention dimension |
|---|---|---|---|---|
| SAGEConv | 3 | 32 | 16 | 8 |
| GraphConv | 3 | 16 | 32 | 16 |
| GCNConv | 3 | 16 | 16 | 8 |
| GATConv | 2 | 16 | 32 | 8 |
| DMPNN | 3 | 32 | 16 | 16 |

**VIRTUAL SCREENING**

Table S3. The candidates predicted to have the highest conductivity in virtual screening.

| Solvent 1 | Solvent 2 | Salt (1 mol/kg) | Log conductivity (S/cm) |
|---|---|---|---|
| C1=CC=CC=C1 | COCOC | F[P-](F)(F)(F)(F)F.[Li+] | -0.7232 |
| ClCCl | FC(F)(C1=NC(C#N)=C([N-]1)C#N)F.CCCCN2C=C[N+](C)=C2 | F[P-](F)(F)(F)(F)F.[Li+] | -0.8839 |
| COCCOC | COCOC | F[P-](F)(F)(F)(F)F.[Li+] | -0.9663 |
| CC1=CC=CC=C1 | COCOC | F[P-](F)(F)(F)(F)F.[Li+] | -1.0576 |
| CC1CCCO1 | COCOC | F[P-](F)(F)(F)(F)F.[Li+] | -1.0785 |
| CC1CCCO1 | COCCOC | F[P-](F)(F)(F)(F)F.[Li+] | -1.1657 |
| C1CCOC1 | COCOC | F[P-](F)(F)(F)(F)F.[Li+] | -1.2042 |
| C1=CC=CC=C1 | COCCOC | F[P-](F)(F)(F)(F)F.[Li+] | -1.2739 |
| C1CCOC1 | CC1CCCO1 | F[P-](F)(F)(F)(F)F.[Li+] | -1.3009 |
| COCOC | FC(F)(C1=NC(C#N)=C([N-]1)C#N)F.CCCCN2C=C[N+](C)=C2 | F[P-](F)(F)(F)(F)F.[Li+] | -1.3355 |

Table S4. Predicted conductivity for mixtures of top candidates' molecular constituents with 1 mol/kg LiPF$_6$.

| Solvent | Salt (1 mol/kg) | Log conductivity (S/cm) |
|---|---|---|
| COCCOC | F[P-](F)(F)(F)(F)F.[Li+] | -1.1057 |
| C1=CC=CC=C1 | | -1.5016 |
| ClCCl | | -1.3809 |
| COCOC | | -0.9134 |
| FC(F)(C1=NC(C#N)=C([N-]1)C#N)F.CCCCN2C=C[N+](C)=C2 | | -1.0740 |
| CC1CCCO1 | | -1.2376 |
| C1CCOC1 | | -1.6093 |